\documentclass[11pt,a4paper]{article}
\usepackage{ACL2021}
\usepackage{times}
\usepackage{latexsym}

\usepackage{microtype}

\usepackage{graphicx}
\usepackage{amsmath}
\usepackage{amsfonts}
\usepackage{booktabs}
\usepackage{multirow}
\usepackage{bm}
\usepackage[english]{babel}

\usepackage{url}
\usepackage{array}

\usepackage{caption}
\usepackage{color} 
\usepackage{makecell}

\newcommand{\PreserveBackslash}[1]{\let\temp=\\#1\let\\=\temp}
\newcolumntype{C}[1]{>{\PreserveBackslash\centering}p{#1}}
\newcolumntype{R}[1]{>{\PreserveBackslash\raggedleft}p{#1}}
\newcolumntype{L}[1]{>{\PreserveBackslash\raggedright}p{#1}}

\title{A Graph Enhanced BERT Model for Event Prediction}

\author {\textbf{Li Du, Xiao Ding\thanks{Corresponding author}, Yue Zhang, Kai Xiong, Ting Liu, and Bing Qin} \\
        Research Center for Social Computing and Information Retrieval \\
        Harbin Institute of Technology, China \\
        \{ldu, xding, kxiong, tliu,qinb\}@ir.hit.edu.cn
        }

\date{}

\begin{document}
\maketitle
\begin{abstract}


Predicting the subsequent event for an existing event context is an important but challenging task, as it requires understanding the underlying relationship between events. Previous methods propose to retrieve relational features from event graph to enhance the modeling of event correlation. However, the sparsity of event graph may restrict the acquisition of relevant graph information, and hence influence the model performance. To address this issue, we consider automatically building of event graph using a BERT model. To this end, we incorporate an additional structured variable into BERT to learn to predict the event connections in the training process.
Hence, in the test process, the connection relationship for unseen events can be predicted by the structured variable.
Results on two event prediction tasks: script event prediction and story ending prediction, show that our approach can outperform state-of-the-art baseline methods. 

\end{abstract}

\section{Introduction}

Understanding the semantics of events and their underlying connections is a long-standing task in natural language processing \cite{minsky1974framework,schank1975scripts}. Much research has been done on extracting script knowledge from narrative texts, and making use of such knowledge for predicting a likely subsequent event given a set of context events. 

A key issue to fulfilling such tasks is the modeling of event relation information. To this end, early work exploited event pair relations \cite{chambers2008unsupervised,jans2012skip,granroth2016happens} and temporal information \cite{pichotta2016using,pichotta2016statistical}. The former has been used for event prediction by using embedding methods, where the similarity between subsequent events and context events are measured and used for candidate ranking. The latter has been used for neural network methods, where models such as LSTMs have been used to model a chain of context events. There has also been work integrating the two methods \cite{wang2017integrating}. 

\begin{figure}[t]
    \centering
    \includegraphics[width=0.9\linewidth]{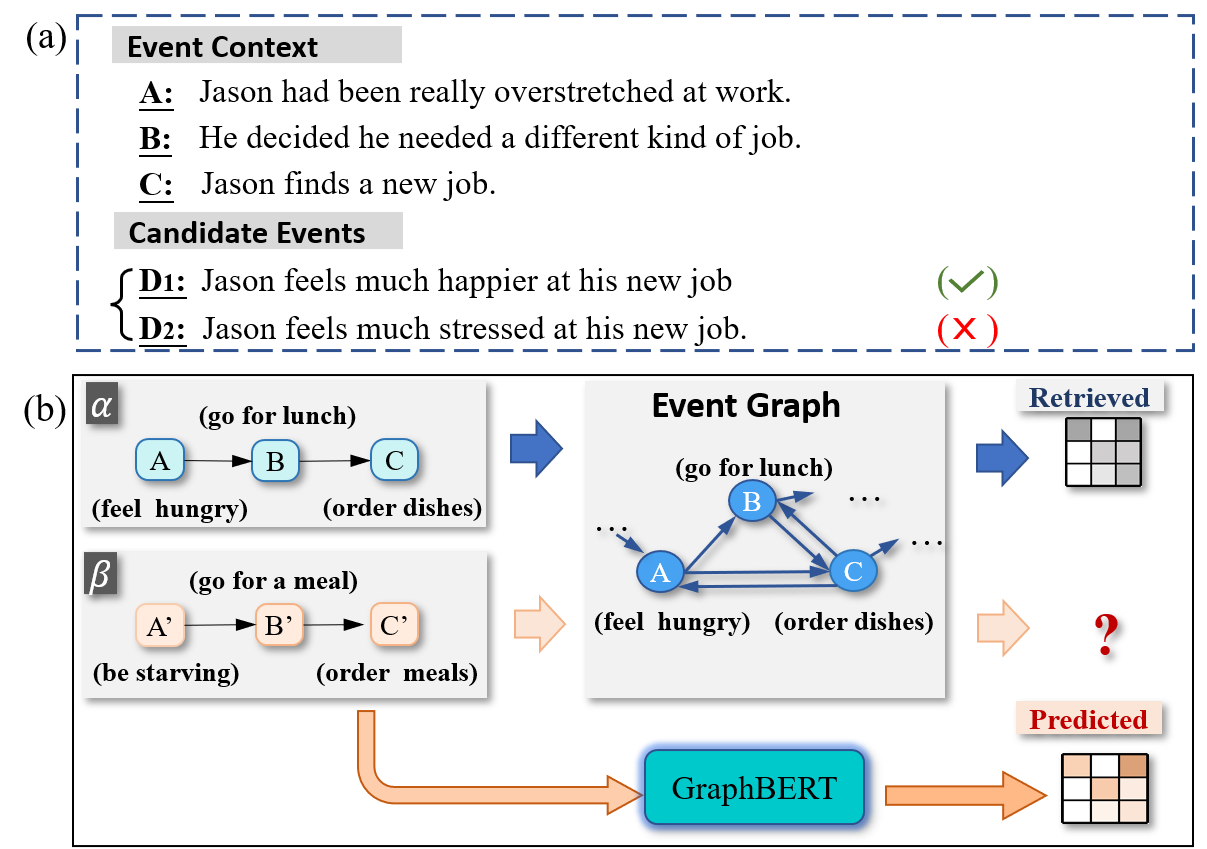}
    \caption{(a) An example for event prediction. (b) Given an event sequence, retrieval-based methods lookup structural information of events from event graph. However, in the test process, part of events may be not covered by the event graph, hence their connection information is unavailable. Different from retrieval-based methods, GraphBERT is able to predict the connection strength between events.}
    \label{fig:example}
\end{figure}

Despite achieving certain effectiveness, the above methods do not fully model the underlying connection between context events. As shown in Figure~1~(a), given the facts that \emph{Jason had been overstretched at work}, \emph{He decided to change job} and \emph{Jason finds a new job}, the subsequent event \emph{Jason is satisfied with his new job} is more likely than \emph{Jason feels much stressed at his new job}, which can be inferred by understanding the fact that the reason for his new job search is stress in his job. \citet{li2018constructing} and \citet{koncel2019text} consider such context structure by building event evolutionary graphs, and using network embedding models to extract relational features. For these methods, event graphs serve as a source of external structured knowledge, which are extracted from narrative texts and provide prior features for event correlation.

One limitation of their methods is that the effectiveness of their methods heavily relies on the coverage of the event graph. As shown in Figure 1 (b), \citet{li2018constructing} and \citet{koncel2019text}'s methods work by looking up the event tuples in the event graph to 
\emph{retrieve} the connection information between events for predicting the output. This is done by the standard knowledge graph lookup operation. However, if the context events are not in the event graph, the method cannot find relevant information. Figure 1 (b) shows an extreme case. In event sequence $\beta$, although the context events \emph{be starving} and \emph{go for a meal} are highly similar to the event graph content \emph{feel hungry} and \emph{go for lunch}, the retrieval-based methods can fail to match context events in the event graph and utilize the event graph knowledge.
However, in practice, it is infeasible to construct an event graph that covers most of the possible events. As an event is the composition of multiple arguments, so the same event can correspond to various semantically equivalent expressions, such as ``feel hungry'' vs ``be starving'', or ``hunger'', etc.
This would limit the performance of the retrieval-based systems. 


To address this issue, we consider automatically predicting the event links using a graph-enhanced BERT model (\textbf{GraphBERT}). As shown in Figure 1 (b), we collect event structure information into a BERT model with graph structure extension. Given a set of event contexts, we use the GraphBERT model to construct an event graph structure by predicting connection strengths between context events, instead of retrieving them from a prebuilt event graph. Specifically, we extend the BERT model by introducing a structured variable, which captures the connection strengths between events.
As shown in Figure 2, during training, both context events and external event graph information are used to train the structured variable. During testing, the structured variable which describes connection strengths between events is obtained using the context event only, which is used for finding the next event. Subsequently, we encode the predicted link strength for making a prediction. 

Experimental results on standard datasets show that our model outperforms baseline methods. Further analysis demonstrates that GraphBERT can predict the connection strengths for unseen events and improve the prediction accuracy. The codes are publicly available at \url{https://github.com/sjcfr}.

\section{Background}
As shown in Figure~\ref{fig:example}~(a), the task of event prediction \cite{mostafazadeh2016corpus,li2018constructing} can be defined as choosing the most reasonable subsequent event for an existing event context. Formally, given a candidate event sequence \small$X=\{X_{e_1},\dots, X_{e_t}, X_{e_{c_j}}\}$\normalsize, where \small$\{X_{e_1},\dots, X_{e_t}\}$\normalsize are $t$ context events and \small$X_{e_{c_j}}$\normalsize is the $c_j$th candidate subsequent event, the prediction model is required to predict a relatedness score $Y \in [0,1]$ for the candidate subsequent event given the event context. 


Event graphs \cite{li2018constructing} have been used to represent relationships between multiple events.
Formally, an event graph could be denoted as $G=\{V, R\}$, where $V$ is the node set, $R$ is the edge set. Each node $V_i\in V$ corresponds to an event $X_i$, while each edge $R_{ij} \in R$ denotes a directed edge $V_i \rightarrow V_j$ along with a weight $W_{ij}$, which is calculated by:
\begin{equation}
\small
  W_{ij}=\frac{\mathrm{count}(V_i, V_j)}{\sum_k \mathrm{count}(V_i, V_k)}
\end{equation}
where $\mathrm{count}(V_i, V_j)$ denotes the frequency of a bigram $(V_i, V_j)$. Hence, the weight $W_{ij}$ is the probability that $X_j$ is the subsequent event of $X_i$. 

\begin{figure*}
    \centering
    \setlength{\abovecaptionskip}{0.1cm}
    \setlength{\belowcaptionskip}{-0.1cm}
    \includegraphics[width=0.9\linewidth]{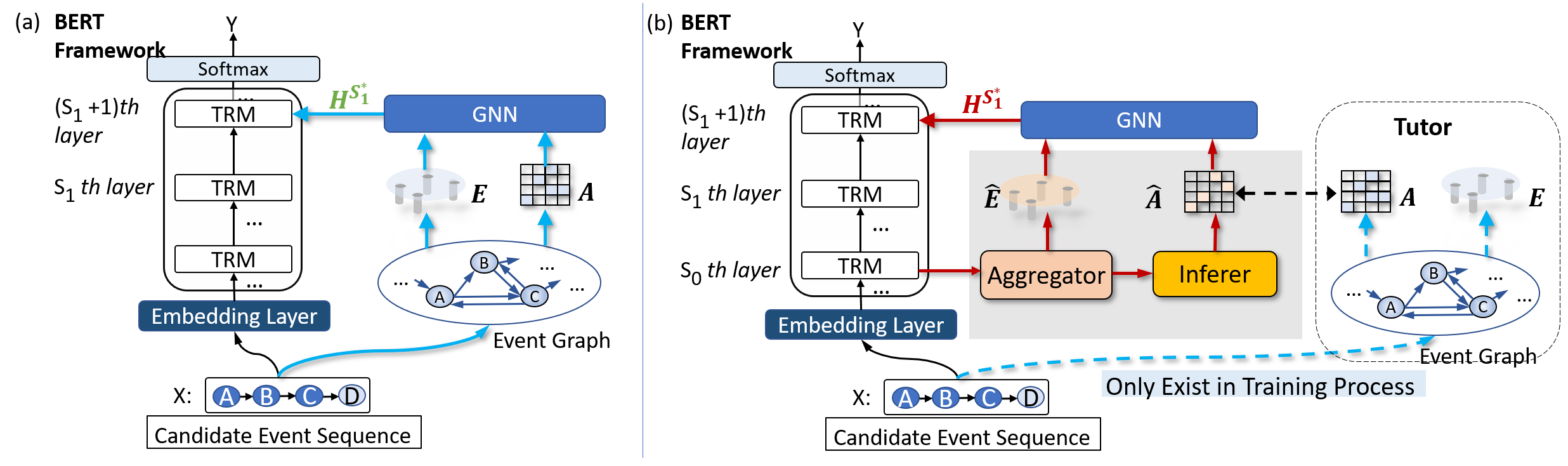}
    \caption{Model Structure. (a) Architecture of the baseline system. Given an event sequence, the baseline system retrieves event node features and connection strength from a prebulit event graph. (b) In addition to the baseline system, GraphBERT introduces an additional \textbf{aggregator} to obtain event representation from the hidden states of BERT, and learns to predict the connection strength between events in the training process using the \textbf{inferer}. So that in the test process, the connection information can be predicted for arbitrary event.}
    \label{fig:model_detail}
\end{figure*}

\section{Baseline System}

Before formally introducing the GraphBERT framework, we first introduce a retrieval-based baseline system. As Figure~\ref{fig:model_detail}~(a) shows, given an event sequence \small$X=\{X_{e_1},\dots, X_{e_t}, X_{e_{c_j}}\}$\normalsize, the baseline system retrieves the corresponding structural information for each event within $X$ from a prebuilt event graph $G$, and then integrates the retrieved structural information into the BERT frame for predicting the relatedness score $Y$. 

For an arbitrary event tuple $(X_{e_i},X_{e_j})$, if it is covered by the event graph $G$ (i.e., both $X_{e_i}$ and $X_{e_j}$ are nodes of $G$), then we can retrieve the corresponding node embeddings $e_i$ and $e_j$, together with the edge weight $A_{ij}$ by matching the event tuple in the event graph. The representation vector of the events within $X$ further form into an embedding matrix $E$, and the edge weights form into an adjacency matrix $A$. 
To make use of the retrieved structural information for enhancing the prediction process, we first employ a graph neural network to combine the event representation matrix and the adjacency matrix:
\vspace{-0.1cm}
\begin{align}
E^{(U)} = \sigma(A E W_U)
\end{align}
where \small $W_U\in \mathbb{R}^{d\times d}$ \normalsize is a weight matrix; $\sigma$ is a sigmoid function; $E^{(U)}$ is the event representation matrix updated by $A$.

Then the combined event graph knowledge can be merged into the frame of BERT for enhancing the prediction process. To this end, we employ an attention operation to softly select relevant information from the updated event representations $E^{(U)}$, and then update the hidden states of BERT. 
Specifically, we take the hidden states of the $s_1$th Transformer layer of BERT (denoted as $H^{s_1}$) as the query, and take the updated event representation $E^{(U)}$ as the key:
\begin{equation}
    {E^{(U)}}^{*}=\mathrm{MultiAttn}(H^{s_1}, E^{(U)})    
\end{equation}
where ${E^{(U)}}^{*}$ carries information selected from $E^{(U)}$ and relevant to $H^{s_1}$.

Then we merge ${E^{(U)}}^{*}$ with $H^{s_1}$ through an addition operation, and employ layer normalization to keep gradient stability:
\begin{equation} 
    {H^{s_1}}^{*}=\mathrm{LayerNorm}({E^{(U)}}^{*}+H^{s_1})
\end{equation}

${H^{s_1}}^{*}$ contains both the node feature information and the connection information between events. By taking ${H^{s_1}}^{*}$ as the input of the subsequent $(s_1+1)$th Transformer layers of BERT, the event prediction process is enhanced with the predicted event graph knowledge.

This retrieval-based baseline system can be regarded as the adaption of \citet{li2018constructing} and \citet{koncel2019text}'s retrieval-based methods on a pretrained model BERT.


\section{GraphBERT}

A critical weakness of the retrieval-based baseline system is that it heavily relies on the coverage of the event graph. In other words, if an event is not covered by the event graph, then the structural information (i.e., node features and the adjacency matrix) would be absent from the constructed event graph, which further limits the model performance.

In this paper, we propose a predictive-based framework GraphBERT. GraphBERT uses the transformer layers of BERT as an encoder to obtain the representation for arbitrary events, and then learns to predict the link strength between events in the training process, so that the sparsity issues in the retrieval process can be avoided.

To this end, as Figure~\ref{fig:model_detail}~(b) shows, in contrast to the retrieval-based baseline system, we introduce two more modules: (1) An \textbf{aggregator} to obtain event representations from the BERT framework; (2) an \textbf{inferer} to predict the link strength between events based on the event representations.

\subsection{Event Encoding}

Given an event sequence $X$, to calculate the event representations and predict the link strength for events within $X$, GraphBERT first encodes $X$ into a set of \emph{token}-level distributed representations by taking the $1$st-$s_0$th Transformer layers of BERT as an encoder. Then an aggregator is employed to aggregate the token level representations into event representations.

\noindent \textbf{Token Level Representations}
For an event sequence \small$X=\{X_1,\cdots,X_{t+1}\}$\normalsize, where \small$X_i=\{x_1,\dots,x_{l_i}\}$\normalsize is an event within $X$ and with $l_i$ tokens, the $s_0$th Transformer layer of BERT encodes these tokens into contextualized distributed representations \small$H^{s_0}=\{(h_1^1,\dots,h^1_{l_{1}}), \cdots, (h_1^{t+1},\dots,h_{l_{t+1}}^{t+1})\}$\normalsize, where $h_j^i \in \mathbb{R}^{1\times d}$ is the distributed representation of the $j$th token of event $X_i$. Then we conduct the graph information prediction as well as the prediction task based on the token representations. 


\noindent \textbf{Event Level Representations}
An \textbf{aggregator module} aggregates tokens representation of events derived from the hidden states of BERT (i.e., $H^{s_0}$) to obtain the event level representations. 
For an arbitrary event $X_i \in X$, we employ a multi-head attention operation \cite{vaswani2017attention} to aggregate information from the corresponding token representations $H_i^{s_0} = (h_1^{i},\dots,h_{l_{i}}^{i})$ and obtain the vector representation of $X_i$. Specifically, we define the query matrix of attention operation as $q_{i}=\frac{1}{l_i}\sum h_{l}^{i}$, and take $H_i^{s_0}$ as the key matrix as well as the value matrix. Then the representation of $X_i$ is calculated as: 
\begin{equation}
  \hat{e}_i=\mathrm{MultiAttn}(q_i, H_i^{s_0}, H_i^{s_0})  
\end{equation}
where $\hat{e}_i \in \mathbb{R}^{1\times d}$.

In this way, we can obtain the representation of all events within $X$, which we denote as \small $\hat{E}=\{\hat{e}_1,\cdots, \hat{e}_{t+1}\}$\normalsize, where \small$\hat{E}\in \mathbb{R}^{(t+1) \times d}$ \normalsize  is a matrix. Note that through the embedding layer of BERT, position information has been injected into the token representations. Thus $\hat{E}$ carries event order information. 

Then the event representation matrix $\hat{E}$ is used for predicting the link strength between events. Hence, the performance of link strength prediction can be strongly influenced by the quality of $\hat{E}$. By deriving $\hat{E}$ from the hidden states of BERT, the abundant language knowledge within BERT can be utilized to obtain the event representations.



\subsection{Link Strength Prediction}

Given the event representation matrix $\hat{E}$ as node features, we employ an \textbf{inferer module} to predict the connection strength between arbitrary events within $X$, regardless of whether these events are seen in the training process. The output is a matrix \small $\hat{A}\in \mathbb{R}^{(t+1)\times (t+1)}$ \normalsize, where $\hat{A}_{ij}$ models the probability that event $j$ is the subsequent event of event $i$.



We stack $n$ graph attention (GAT) layers \cite{velivckovic2017graph} for consolidating event features. For an event $X_i$, the GAT layer works on the neighborhood of $X_i$ to aggregate information. Since the connection between events are unknown a priori, we set the neighborhood set of event $X_i$ as $\mathcal{N}_i=\{X_j\}$, where $X_j \in X, j \neq i$.  

Therefore, at the $k$th graph attention layer, given the representation of the $i$th event $\hat{e}^{k}_i$, we calculate the attention coefficients between other events and derive deep event representation as:
\vspace{-0.1cm}
\begin{equation}
\small
\begin{split}
    \alpha_{ij}=\mathrm{softmax}_{j, j\in \mathcal{N}_i} &(\mathrm{Relu}(u[W_{\alpha}\hat{e}^{k}_i||W_{\alpha}\hat{e}^{k}_j])) \\
    \hat{e}^{k+1}_i= \sigma &(\sum_{j \in \mathcal{N}_{i}} \alpha_{i j} \mathbf{W}_{\alpha} \hat{e}^{k}_j)
\end{split}
\end{equation}
where \small$u \in \mathbb{R}^{1\times 2d },  W_{\alpha} \in \mathbb{R}^{d \times d}$\normalsize are trainable parameters, \small$\cdot || \cdot$\normalsize is a concatenation operation. At the first GAT layer, \small$\hat{e}^{1}_i$\normalsize is initialized by \small$\hat{e}_i$\normalsize derived from the aggregator.

After $n$ graph attention operations, we employ a bilinear map to calculate a relation strength score between two events within $X$ based on their deep representations:
\begin{equation}
\Gamma_{ij}=\left( \hat{e}^{n}_i \ W_R \ \mathrm{T}({\hat{e}^{n}_j})\right)
\end{equation}
where $W_R\in \mathbb{R}^{d\times d}$ are learnable parameters, $T(\cdot)$ is the transpose operation. For all $t+1$ events within $X$, the relation strength score between arbitrary two events forms a matrix \small$\Gamma \in \mathbb{R}^{(t+1)\times (t+1)}$\normalsize, with each element $\Gamma_{ij}$ measuring the relation strength between $X_i$ and $X_j$.

Then we normalize the relation strength scores using the softmax function:
\begin{equation}
\hat{A}_{ij}=\mathrm{softmax}_j(\Gamma_{ij})
\end{equation}
After the layer normalization, $\sum_j \hat{A}_{ij}=1$. 

Hence, with the aggregator and the inferer, GraphBERT can obtain representation and connection strengths for arbitrary events, regardless of whether or not the event is covered by the event graph. Then the predicted adjacency matrix $\hat{A}$ and event representations $\hat{E}$ can be used for prediction, and the process is same as the retrieval-based baseline, as described in Eq.(2)-Eq.(4).


\subsection{Training of Inferer}
In the training process, we employ a \textbf{tutor module} to supervise the prediction of $\hat{A}$ using the structural information from a prebuilt event graph. 
Given an event sequence $X$, the tutor obtains an adjacency matrix $A$ based on the edge weights of the event graph. 
Formally, the weights of $A$ are initialized as:
\begin{equation}
\small
A_{ij}=\left\{
\begin{aligned}
&W_{ij}, & \mathrm{if} \, V_{i^{'}} \rightarrow V_{j^{'}} \in R, \\
&0, & \mathrm{others}.
\end{aligned}
\right.
\end{equation}
\noindent where \small$V_{i^{'}}$, $V_{j^{'}}$\normalsize are nodes in the event graph corresponding to the $i$th and the $j$th event of the candidate event sequence. The same as the predicted event adjacency matrix \small$\hat{A}$\normalsize, \small$A$ \normalsize is also a \small$\mathbb{R}^{(t+1)\times (t+1)}$\normalsize matrix.

We scale $A$ to make each row sum equals 1. Therefore, each element of $A$ models the probability that the $j$th event is the subsequent event of the $i$th event in $X$. In the training process, through minimizing the distance between $\hat{A}$ and $A$, the inferer module is supervised by the tutor to learn to predict the event connection strength based on the event representations.  



\subsection{Optimization}
The overall loss function is defined as:
\vspace{-0.2cm}
\begin{align}
\small
  \begin{split}
    &L=L_{\text{Event Prediction}}+\lambda L_{\text{Graph Reconstruction}}\\
  \end{split}
\label{eq:2}
\end{align}
where $L_{\text{Event Prediction}}$ is a cross-entropy loss measuring the difference between predicted relatedness score $Y$ and golden label, $L_{\text{Graph Reconstruction}}$ assess the difference between $A$ and $\hat{A}$, $\lambda$ is an additional hyperparameter for balancing the prediction loss with graph reconstruction loss. 

For calculating $L_{\text{Graph Reconstruction}}$, 
we cast both $A$ and $\hat{A}$ as a set of random variables, and employ the KL divergence to measure their difference: 
\begin{align}
\small
\begin{split}
\vspace{-0.2cm}
&L_{\text{Graph Reconstruction}}=\\
\vspace{-0.2cm}
&\sum_i \mathrm{KL}(\mathrm{MultiNomial}(\hat{A}_i)||\mathrm{MultiNomial}(A_i))
\end{split}
\end{align}

where $i$ denotes the $i$th row, and $\mathrm{MultiNomial}(\cdot)$ denotes the multinomial distribution.



\section{Experiments}


We evaluate our approach on two event prediction tasks: Multiple Choice Narrative Cloze Task (MCNC) \cite{granroth2016happens} and Story Cloze Test (SCT) \cite{mostafazadeh2016corpus} by constructing an event graph based on the training set of MCNC to train the GraphBERT model and then adapts the GraphBERT model trained on the MCNC dataset to the SCT dataset to evaluate whether GraphBERT can predict the link strength between unseen events to enhance the prediction performance.

\subsection{Dataset} 

\noindent \textbf{Multiple Choice Narrative Cloze Task}
The MCNC task requires the prediction model to choose the most reasonable subsequent event from five candidate events given an event context \cite{granroth2016happens}. In this task, each event is abstracted to Predicate-GR form \cite{granroth2016happens}, which represents an event in a structure of \{subject, predicate, object, prepositional object\}. 
Following \citeauthor{granroth2016happens} (\citeyear{granroth2016happens}), we extract event chains from the New York Times portion of the Gigaword corpus. 
The detailed statistics of the dataset are shown in Table~\ref{tab:stat_mcnc}. 

\noindent \textbf{Story Cloze Test Task}
The SCT task requires models to select the correct ending from two candidates given a story context. Compared with MCNC which focuses on abstract events, the stories in SCT are concrete events and with much more details. 
This dataset contains a five-sentence story training set with 98,162 instances, and 1,871 four-sentence story contexts along with a right ending and a wrong ending in the dev. and test dataset, respectively. Because of the absence of wrong ending in the training set, we only use the development and the test dataset, and split the development set into 1,771 instances for finetuning models and 100 instances for the development purpose. 

\subsection{Construction of Event Graph}
\begin{table}
    \setlength{\abovecaptionskip}{0.09cm}
    \setlength{\belowcaptionskip}{-0.1cm}
    \centering
    \small
    \begin{tabular}{lccc}
    \hline
       & \textbf{Training} & \textbf{Dev.} & \textbf{Test} \\
    \hline
    \#Documents & 830,643 & 103,583 & 103,805 \\
    \#Event Chains & 140,331 & 10,000 & 10,000  \\
    \hline
    \#Unique Events & 430,516 & 44,581 & 47,252 \\
    \#Uncovered Events & 0 & 24,358 & 24,081 \\
    \hline
    \end{tabular}
    \caption{Statistics of the MCNC dataset.}
    \vspace{-0.5cm}
    \label{tab:stat_mcnc}
\end{table}
The event graph is constructed based on the training set of the MCNC dataset. Each event within the training set of MCNC is taken as a node of the event graph, and the edge weights are obtained by calculating the event bigram frequency.
Note that, as shown in Table~\ref{tab:stat_mcnc}, although the events have been processed into a highly abstracted form to alleviate the sparsity, there are still nearly half of the events in the development and test set of MCNC remains uncovered by the event graph. In the test process, for retrieval-based methods, given a candidate event sequence with length $t+1$, the edge weights for events not covered by the event graph are all set as $1/(t+1)$.

\subsection{Experimental Settings}

We implement the GraphBERT model using pretrained BERT-base model, which contains 12 Transformer layers. We aggregate the \emph{token} representations from the 7th Transformer layer of BERT, and merge the updated event representations to the 10th Transformer layer of BERT. The aggregator has a dimension of 768, and contains 12 attention heads. The inferer contains 1 GAT layer. The balance coefficient $\lambda$ equals 0.01. During the training and testing process, we concatenate the elements of the Predicate-GRs to turn the Predicate-GRs into strings, so that the event sequences can conform to the input format of the GraphBERT model. More details are provided in the Appendix.

\noindent \textbf{Baselines for MCNC} 

\noindent \textbf{Event Pair and Event Chain Based Methods}

\textbf{(i) Event-Comp} \cite{granroth2016happens}  calculates the pair-wise event relatedness score using a Siamese network.
\textbf{(ii) PairLSTM}  \cite{wang2017integrating} integrates event order information and pairwise event relations to predict the ending event.
\textbf{(ii) RoBERTa-RF} \cite{lv2020integrating} enhances pretrained language model RoBERTa with chain-wise event relation knowledge for making prediction.

\noindent \textbf{Event Graph Based Methods}

\textbf{(i) SGNN} \cite{li2018constructing} constructs a narrative event evolutionary graph (NEEG) to describe event connections, and propose a scaled graph neural network to predict the ending event based on structural information \textbf{\emph{retrieved}} from the NEEG.
\textbf{(ii) HeterEvent} \cite{zheng2020heterogeneous} encodes events using BERT, and implicitly models the word-event relationship by an heterogeneous graph attention mechanism.
\textbf{(iii) GraphTransformer} \cite{koncel2019text} \textbf{\emph{retrieves}} structural information from event graph and introduces an additional graph encoder upon BERT to leverage the structural information.

\noindent \textbf{Pretrained Language Model Based Methods}

\textbf{(i) BERT} \cite{devlin2019bert} refers to the BERT-base model finetuned on the MCNC dataset. 
\textbf{(ii) GraphBERT$_{\lambda=0}$} refers the GraphBERT model optimized with the balance coefficient $\lambda$ set as 0. Hence, the structural information cannot be incorporated through the graph reconstruction term.

\subsubsection{Settings for SCT} 

To test the generality of GraphBERT, we examine whether GraphBERT can utilize the structural knowledge learned from MCNC-based event graph to guide the SCT task.
To make fair comparisons, we also trained the BERT \cite{devlin2019bert}, GraphTransformer \cite{koncel2019text} on the MCNC dataset, then finetuned them on the SCT dataset. In the following sections, we use the subscript ``MCNC'' to denote the model which has been trained on the MCNC dataset.

However, in the finetuning and test process, GraphTransformer still relies on an event graph to provide structural information. To address this issue, we abstract each event in the finetuning set and test set of SCT into the Predicate-GR form, which is the same form with the nodes in the MCNC-based event graph. As a result, structural information for an event in SCT can be retrieved from the MCNC-based event graph using its corresponding Predicate-GR form, once the event is covered by the event graph.

In addition to the above-mentioned methods, on the SCT dataset, we also compare GraphBERT with the following \textbf{event-chain-based} baselines:

\textbf{(i) HCM} \cite{chaturvedi2017story} trains a logistic regression model based on contextual semantic features.
\textbf{(ii) ISCK} \cite{chen2019incorporating} integrates narrative sequence and sentimental evolution information to predict the story ending.

\subsubsection{Overall Results} 

We list the results on MCNC and SCT in Table~\ref{tab:res_mcnc} and Table~\ref{tab:res_sct}, respectively. From the results on MCNC (Table~\ref{tab:res_mcnc}), we can observe that:

(1) Compared to event-pair-based EventComp and event-chain-based PairLSTM, event-graph-based methods (i.e. SGNN, 
HeterEvent, GraphTransformer, and GraphBERT) show better performance.  In addition, GraphBERT outperforms event-chain based RoBERTa-RF, though RoBERTa-RF is built upon a much more powerful language model. This confirms that involving event structural information could be effective for this task.


(2) Compared to BERT and GraphBERT$_{\lambda=0}$, graph enhanced models GraphTransformer and GraphBERT further improve the accuracy of script event prediction (T-test; P-Value $<$ 0.01). This shows that linguistic and structural knowledge can have a complementary effect. 


(3) Compared to the retrieval-based method GraphTransformer, GraphBERT shows efficiency of learning structural information from the event graph (T-test; P-Value $<$ 0.01). This indicates that GraphBERT is able to learn the structural information from the event graph in the training process, and predict the correct structural information for unseen events in the test process.

Results on the SCT dataset (Table~\ref{tab:res_sct}) show that:


(1) Comparing GraphBERT with $\text{BERT}_{\text{MCNC}}$, $\text{GraphBERT}_{\lambda=0, \text{MCNC}}$ shows that the graph information can also be helpful for the SCT task.

(2) Though incorporated graph information, the performance of GraphTransformer is close or inferior to BERT on SCT. This could be because of the limited size of the SCT development set, which contains 1,771 samples and might be insufficient to adapt GraphTransformer to the SCT problem. However, GraphBERT shows a 1.3\% absolute improvement over BERT, which indicates the efficiency of GraphBERT in predicting the link strength between unseen events for predicting the ending event.

\begin{table}[]
    \small
    \centering
    \setlength{\abovecaptionskip}{0.1cm}
    \setlength{\belowcaptionskip}{-0.31cm}
    \begin{tabular}{lc}
    \hline
    \textbf{Methods} & \textbf{Accuracy(\%)} \\
    \hline
    Random  & 20.00**   \\
    EventComp \cite{granroth2016happens} & 49.57** \\
    PairLSTM \cite{wang2017integrating} & 50.83** \\
    SGNN \cite{li2018constructing} & 52.45** \\
    BERT \cite{devlin2019bert} & 57.35** \\
    GraphTransformer \cite{koncel2019text}  & 58.53** \\
    HeterEvent \cite{zheng2020heterogeneous} & 58.10** \\
    GraphBERT$_{\lambda=0}$ &  57.23** \\
    RoBERTa-RF \cite{lv2020integrating} & 58.66** \\
    \hline
    GraphBERT & \textbf{60.72} \ \ \ \ \\
    \hline
    \end{tabular}
    \caption{Performance of GraphBERT and baseline methods on the test set of MCNC. Accuracy marked with * means
p-value $<$ 0.05 and ** indicates p-value $<$ 0.01 in T-test.}
    \label{tab:res_mcnc}
\end{table}

\begin{table}[]
    \small
    \centering
    \setlength{\belowcaptionskip}{-0.31cm}
    \begin{tabular}{lc}
    \hline
    \textbf{Methods} & \textbf{Accuracy(\%)} \\
    \hline
    HCM \cite{chaturvedi2017story} & 77.6** \\
    ISCK \cite{chen2019incorporating} & 87.6** \\
    BERT \cite{devlin2019bert} & 88.1* \ \ \\
    $\text{BERT}_{\text{MCNC}}$ & 88.5* \ \ \\
    $\text{GraphTransformer}_{\text{MCNC}}$ \scriptsize{\cite{koncel2019text}}  & 88.9 \ \  \ \ \\
    HeterEvent$_\text{MCNC}$ \cite{zheng2020heterogeneous} & 88.4* \ \ \ \\
    $\text{GraphBERT}_{\lambda=0, \text{MCNC}}$ & 88.3* \ \ \\
    \hline
    $\text{GraphBERT}_{\text{MCNC}}$ & \textbf{89.8} \ \ \ \ \\
    \hline
    \end{tabular}
    \caption{Model performance on the test set of SCT. Accuracy marked with * means
p-value $<$ 0.05 and ** indicates p-value $<$ 0.01 in T-test.}
    \label{tab:res_sct}
\end{table}

\subsection{Influence of the Accuracy of the Predicted Link Strength}

\begin{figure}[t]
    \centering
    \setlength{\abovecaptionskip}{0.1cm}
    \setlength{\belowcaptionskip}{-0.21cm} 
    \includegraphics[width=0.85\linewidth]{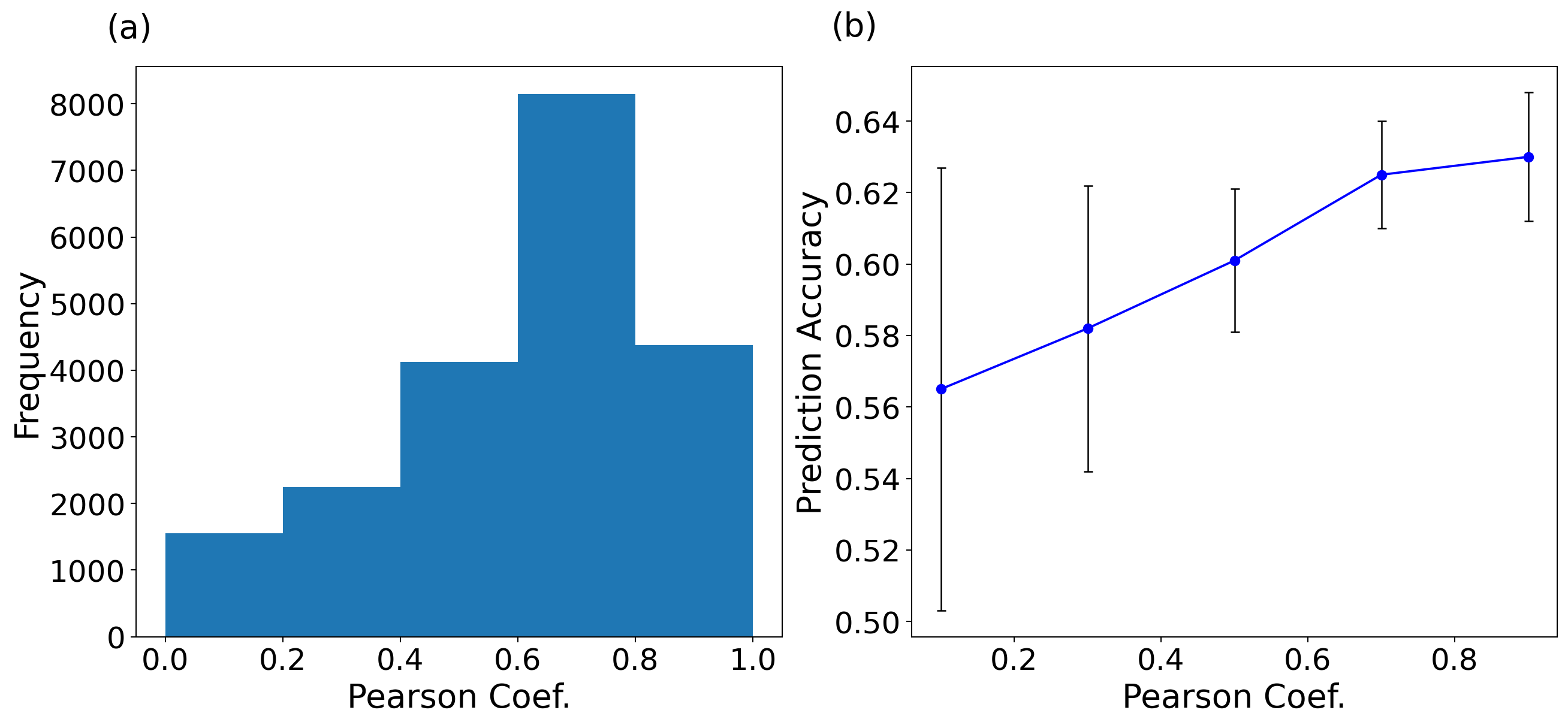}
    \caption{(a) The distribution of Pearson correlation coefficients between the predicted connection strength and connection strength derived from the event graph. (b) Relationship between correlation coefficient and model performance.}
    \label{fig:qua}
\end{figure}

We investigate the relationship between the accuracy of the predicted link strengths with the model performance. 
However, for events in the test set, the golden event graph is unavailable. To address this issue, we split the original training set of MCNC into a new training and evaluating set, containing 120,331 and 20,000 instances, respectively. 
For each sample, we calculate the Pearson correlation coefficient between the predicted connection strengths and connection strengths derived from the event graph, as well as the relationship between such correlation coefficient and model performance. The results are shown in Figure~\ref{fig:qua}. We observe that, in general, GraphBERT can predict the connection between arbitrary events with reasonable accuracy. Also, the model performance improves as the connection prediction accuracy increases. This confirms that correctly predicting the event connections for unseen events can be helpful for the event prediction process.

\subsection{Influence of the Coverage of the Event Graph}

\begin{figure}
    \setlength{\abovecaptionskip}{0.01cm}
    \setlength{\belowcaptionskip}{-0.21cm}
    \centering
    \includegraphics[width=0.7\linewidth]{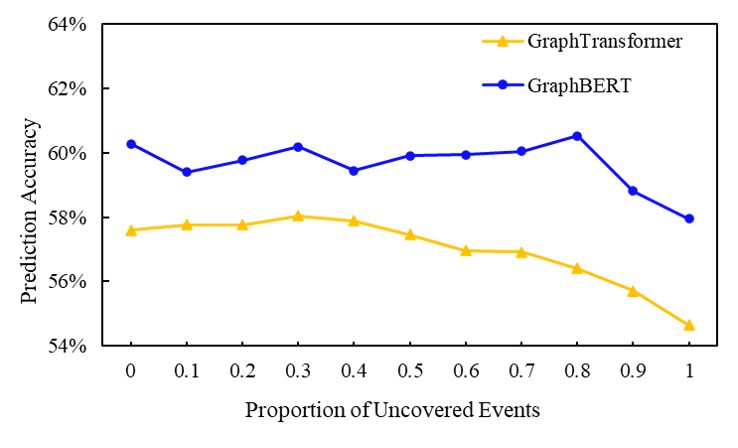}
    \caption{The performance of GraphBERT and GraphTransformer under different proportion of uncovered events.}
    \label{fig:accu_uncov}
\end{figure}

We conduct experiments to investigate the specific influence of the sparsity of the event graph on model performance. Based on the original test set of MCNC, we build new test sets with different proportions of uncovered events, and compare the performances of the GraphBERT framework with retrieval-based method GraphTransformer \cite{koncel2019text} on these test sets. 
As shown in Figure~\ref{fig:accu_uncov}, as the proportion of uncovered events increase from 0 to 1, the performance of GraphTransformer shows a negative trend in general. This is because, for retrieval-based methods, with the increase of sparsity, the availability of structural information decreases. Compared to GraphTransformer, the performance of GraphBERT is more stable. These results indicate that predicting the structural information can be useful for enhancing the performance of event prediction.

\begin{table*}[ht]
\small
\centering
\setlength{\abovecaptionskip}{0.2cm}
\setlength{\belowcaptionskip}{-0.21cm}
\begin{tabular}{c|c|c}
\hline
Event Context & Candidate Subsequent Event & Model  \\
\hline
\multirow{3}*{  \makecell*[l]{\textbf{A:} I heard that my school's campus had been closed. \\ \textbf{B:}  The message said there was a bear on the grounds ! \\ \textbf{C:} The police had to come and help get the bear away. \\
\textbf{D:} They gave the bear a tranquilizer. }  }  & \multirow{2}*{$\bm{\mathrm{E_1}}$:  The bear fell asleep. ($\surd$)} & \multirow{2}*{GraphBERT} \\
~  & ~ & ~ \\
 \cline{2-3}
~ & \multirow{2}*{\makecell*[l]{ $\bm{\mathrm{E_2}}$: The bear became very violent. ($\times$)}} & \multirow{2}*{GraphTransformer} \\
~  & ~ & ~ \\[3pt]

\hline
\end{tabular}
\caption{An example of event predictions made by GraphTransformer and GraphBERT on the SCT dataset.}
\label{tab:case_study}
\end{table*}

\subsection{Case Study}

Table~\ref{tab:case_study} provides an example of prediction results from different models on the test set of SCT. The event context describes a story that a bear appeared in the campus and policemen came to tranquilize the bear. Given the event context, GraphBERT is able to choose correct ending $\mathrm{E_1}$ \emph{The bear fell asleep}, while GraphTransformer chooses the incorrect ending $\mathrm{E_2}$ \emph{The bear became very violent}. 

To correctly predict the story ending, a model should understand the relationship between \emph{gave a tranquilizer} and \emph{fell asleep}. However, event \emph{gave a tranquilizer} is not covered by the event graph. Hence, the retrieval-based method GraphTransformer is unable to obtain structural information from the event graph. On the other hand, in the event graph, there is a directed edge from a node \emph{obj. sedated} to node \emph{subj. slept}. This indicates that, GraphBERT can learn the structural knowledge from the MCNC-based event graph, and predict the connection between \emph{gave a tranquilizer} and \emph{fell asleep} for instances in the SCT dataset. 

\section{Discussion}

The GraphBERT model employs a structure variable $\hat{A}$ to capture the ``$\text{is\_next\_event}$'' relationship between events. By introducing more parallel structural variables \small$\{\hat{A}^{1},\dots, \hat{A}^{k}\}$\normalsize, it can be extended to simultaneously learn multiple kinds of event relationships, such as temporal or causal relationship. Furthermore, previous researches demonstrate that the graph-structured relationship extensively exist between other semantic units, such as sentences\cite{yasunaga2017graph}, or even paragraphs \cite{sonawane2014graph}. However, similar to the situation in event graph, it would be impractical to construct knowledge graphs that cover all possible connection relationships between all the sentences or paragraphs. This restricts the applicable of retrieval-based methods in these situations. On the contrary, our generative approach suggests a potential solution by learning the connection relationship from graph-structured knowledge base with limited size, then generalizing to the unseen cases. 

\section{Related Work}
The investigation of scripts dates back to 1970's \cite{minsky1974framework,schank1975scripts}. The script event prediction task models the relationships between abstract events. Previous studies propose to model the pair-wise relationship \cite{chambers2008unsupervised,jans2012skip,granroth2016happens} or event order information \cite{pichotta2016statistical,pichotta2016using,wang2017integrating} for predicting the subsequent event. \citeauthor{li2018constructing} (\citeyear{li2018constructing}) and \citeauthor{lv2019sam} (\citeyear{lv2019sam}) propose to leverage the rich connection between events using graph neural network and attention mechanism, respectively.

Different from script event prediction, the story cloze task \cite{mostafazadeh2016corpus} focuses on concrete events. Therefore, it requires prediction models to learn commonsense knowledge for understanding the story plot and predicting the ending. To this end, 
\citeauthor{li2018multi} (\citeyear{li2018multi}) and \citeauthor{guan2019story} (\citeyear{guan2019story}) propose to combine context clues with external knowledge such as KGs.  \citeauthor{li2019story} (\citeyear{li2019story}) finetune pretrained language models to solve the task.
Compared to their works, our approach can use both the language knowledge enriched in BERT to promote the comprehension of event context, and the structural information from event graph to enhance the modeling of event connections. 

A recent line of work has been engaged in combining the strength of Transformer based models with graph structured data. To integrate KG with language representation model BERT, \citet{zhang2019ernie} encode KG with a graph embedding algorithm TransE \cite{bordes2013translating}, and takes the representation of entities in KG as input of their model. 
However, this line of work only linearizes KGs to adapt the input of BERT. Graph structure is not substantially integrated with BERT. \citeauthor{guan2019story} (\citeyear{guan2019story}) and \citeauthor{koncel2019text} (\citeyear{koncel2019text}) propose retrieval-based methods to leverage the structural information of KG. However, in the event prediction task, the diversity of event expression challenges the coverage of the event graph, and prevents us from simply retrieving events in the test instances from the event graph.
We propose to integrate the graph structural information with BERT through a predictive method. Compared to retrieval-based methods, our approach is able to learn the structural information of the event graph and \emph{generate} the structural information of events to avoid the unavailable of structural information in test instances.
\section{Conclusion}
We devised a graph knowledge enhanced BERT model for the event prediction task. In addition to the BERT structure, GraphBERT introduces a structured variable to learn structural information from the event graph, and model the relationship between the event context and the candidate subsequent event. Compared to retrieval-based methods, GraphBERT is able to predict the link strength between all events, thus avoiding the (inevitable) sparsity of event graph. Experimental results on MCNC and SCT task show that GraphBERT can improve the event prediction performances compared to state-of-the-art baseline methods. In addition, GraphBERT could also be adapted to other graph-structured data, such as knowledge graphs. 

\section{Acknowledgments}

We thank the anonymous reviewers for their constructive comments, and gratefully acknowledge the support of the New Generation Artificial Intelligence of China (2020AAA0106501), and the National Natural Science Foundation of China (62176079, 61976073).

\bibliography{acl2021}
\bibliographystyle{acl_natbib}

\section{Experimental Settings}
\subsection{Training Details}

To conform to the input format of BERT, for an event described in the Predicate-GR form \{subject, predicate, object, prepositional object\}, we first concatenate each element within the predicate-GR into a string ``subject predicate object prepositional object'', so that an event described in a structured form is turned into a string. Then for satisfying the requirement of BERT, the candidate event sequence is further preprocessed into the form of:
\begin{equation}
\small
\texttt{[CLS]  $e_1$ [SEP] $\dots e_t$ [SEP] candidate [SEP]}
\end{equation}

On the MCNC dataset, the GraphBERT model is trained for 3 epochs, with a batch size of 64, and a learning rate of 2e-5. While during the finetuning process on SCT, GraphBERT is optimized with a batch size of 16, and a learning rate of 1e-5, with 5 epochs.

\subsection{Searching for the Balance Coefficient}

\begin{figure}[t]
    \centering
    \includegraphics[width=0.8\linewidth]{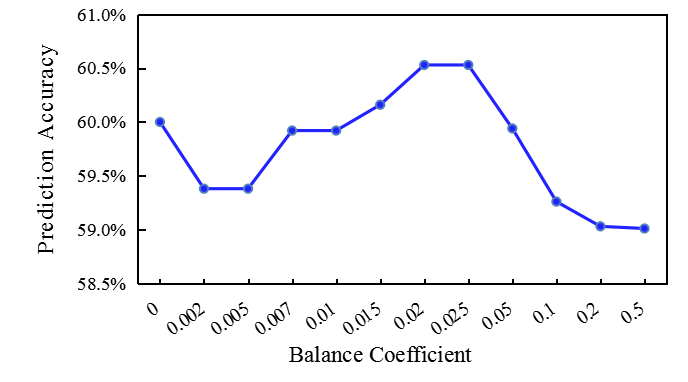}
    \caption{The performance of model trained with different balance coefficient $\lambda$.}
    \label{fig:accu_lambda}
\end{figure}

In this paper, the objective function is composed of two components. Through minimizing the graph reconstruction loss, model learns to modeling the bigram event adjacency patterns. While through minimizing the prediction loss, model is trained to choose the correct ending given an event context. These two components are balanced with a coefficient $\lambda$.

To investigate the effect of the balance coefficient, we compare the prediction accuracy of the GraphBERT model trained with different $\lambda$ and show the results in Figure~\ref{fig:accu_lambda}. From which we could observe that, the prediction accuracy increases as the balance coefficient increase from 0 to 0.1. This is because the additional event graph structure information is helpful for the event prediction task. However, as the $\lambda$ exceeds 0.5, the model performances start to decrease. This is because the overemphasis of graph reconstruction loss would in turn decrease the model performance.

\begin{table}[]
    \small
    \centering
    \begin{tabular}{cccccc}
    \hline
 \textbf{(4, 10)} & \textbf{(5, 10)} & \textbf{(6, 10)} & \textbf{(7, 10)} & \textbf{(8, 10)}& \textbf{(9, 10)} \\
    \hline
    58.76 & 60.28 & 60.57 & \textbf{60.72} & 60.28 & 60.01 \\
    \hline
    \end{tabular}
    \caption{Influence of start layer and merge layer on model performance.}
    \label{tab:ablation}
\end{table}

\subsection{Searching of Start and Merge Layer in BERT}

Different transformer layers of BERT tend to concentrate on different semantic and syntactic information \cite{clark2019does,coenen2019visualizing}.
Therefore, which layer is selected in the BERT to start integrating event graph knowledge, and which layer is selected to merge graph enhanced event representations can affect the performance of the model. We study such effect in two ways: first, we fix the start layer and change the merge layer. Second, we fix the gap between start and merge layer, and change the start layer. Results are shown in Table~\ref{tab:ablation}. The tuple ($n_1$,  $n_2$) denotes the (start, merge) layer. From which we could observe that, under the same gap between merge and start layer, employing the 7th transformer layer of  BERT as the start layer can achieve the best result. While setting the merge--start gap as 2 is more efficient than other choices. Interestingly, \citet{jawahar2019does} find that the syntactic features can be well captured in the middle layers of BERT, especially in the 7--9 layer. This indicates that the middle layers of BERT focus more on sentence level information, and implicitly support the reasonableness that choosing the 7th and 10th transformer layer of BERT as the start end merge layer.

\section{Enhancing Different Kinds of Pretrained Transformer-based Pretrained Language Models with Event Graph Knowledge}

\begin{table}[]
    \small
    \centering
    \begin{tabular}{cc}
    \hline
    \textbf{Model} & \textbf{Prediction Accuracy (\%)} \\
    \hline
    BERT & 57.35\\
    GraphBERT & 60.72 \\
    RoBERTa & 61.19 \\
    GraphRoBERTa & 62.81\\
    \hline
    \end{tabular}
    \caption{Performance of the event graph knowledge enhanced RoBERTa model (Graph-RoBERTa) on the MCNC dataset.}
    \label{tab:rbt}
\end{table}

In this paper, we propose the GraphBERT framework, which enhances the transformer-based pertrained language model BERT with event graph knowledge through an additional structural variable $\hat{A}$. We argue that, using the structural variable, we can also equip other transformer-based pretrained language models, such as RoBERTa, with the event graph knowledge, and then enhance the event prediction process. This could be achieved by adapt the aggregator, inferer and merger module upon the other transformer-based frameworks. 

Using the above-mentioned manner, we implemented a GraphRoBERTa model and examined its performance on the MCNC dataset. The results are shown in Table~\ref{tab:rbt}. We observe that, compared with BERT, RoBERTa and GraphRoBERTa show better performance. This is because, during the pretraining process, RoBERTa can acquire more abundant linguistic knowledge for understanding the events through the dynamic masked token prediction mechanism. Moreover, the comparison between GraphBERT with BERT, and between GraphRoBERTa with RoBERTa show the effectiveness of our approach in incorporating event graph knowledge with multiple prevailing transformer-based pretrained language models, to consistently enhancing the performance of event prediction.

\end{document}